\documentclass[10pt,twocolumn,letterpaper]{article}

\usepackage{iccv}
\usepackage{times}
\usepackage{epsfig}
\usepackage{graphicx}
\usepackage{amsmath}
\usepackage{amssymb}
\usepackage{multirow}
\usepackage{pifont}
\usepackage[ruled,vlined,linesnumbered,ruled]{algorithm2e}

\usepackage[breaklinks=true,bookmarks=false]{hyperref}

\iccvfinalcopy 

\def\iccvPaperID{****} 

\ificcvfinal\fi

\begin{document}

\title{An Efficient Transformer for Simultaneous Learning of BEV and Lane Representations in 3D Lane Detection}


\author{
	Ziye Chen\textsuperscript{1}  \ \ \ \
	Kate Smith-Miles\textsuperscript{1}  \ \ \ \
	Bo Du\textsuperscript{2}  \ \ \ \
	Guoqi Qian\textsuperscript{1}  \ \ \ \
	Mingming Gong\textsuperscript{1} \\
	\textsuperscript{1} School of Mathematics and Statistics, University of Melbourne, Australia \\
	\textsuperscript{2}School of Computer Science, Wuhan University, Wuhan, China\\
	\tt ziyec1@student.unimelb.edu.au, kate.smithmiles@gmail.com, \\
    \tt dubo@whu.edu.cn, qguoqi@unimelb.edu.au, mingming.gong@unimelb.edu.au
}

\maketitle
\ificcvfinal\thispagestyle{empty}\fi

\begin{abstract}

Accurately detecting lane lines in 3D space is crucial for autonomous driving. Existing methods usually first transform image-view features into bird’s-eye-view (BEV) by aid of inverse perspective mapping (IPM), and then detect lane lines based on the BEV features. However, IPM ignores the changes in road height, leading to inaccurate view transformations. Additionally, the two separate stages of the process can cause cumulative errors and increased complexity.
To address these limitations, we propose an efficient transformer for 3D lane detection. Different from the vanilla transformer, our model contains a decomposed cross-attention mechanism to simultaneously learn lane and BEV representations. The mechanism decomposes the cross-attention between image-view and BEV features into the one between image-view and lane features, and the one between lane and BEV features, both of which are supervised with ground-truth lane lines. Our method obtains 2D and 3D lane predictions by applying the lane features to the image-view and BEV features, respectively. This allows for a more accurate view transformation than IPM-based methods, as the view transformation is learned from data with a supervised cross-attention. Additionally, the cross-attention between lane and BEV features enables them to adjust to each other, resulting in more accurate lane detection than the two separate stages. Finally, the decomposed cross-attention is more efficient than the original one.
Experimental results on OpenLane and ONCE-3DLanes demonstrate the state-of-the-art performance of our method.
\end{abstract}

\section{Introduction}
Lane detection is a crucial component of assisted and autonomous driving systems, as it enables a range of downstream tasks such as route planning, lane keeping assist, and high-definition (HD) map construction \cite{li2022hdmapnet}. In recent years, deep-learning-based lane detection algorithms have achieved impressive results in 2D image space. However, in practical applications, it is often necessary to represent lane lines in 3D space or bird's-eye-view (BEV). This is particularly useful for tasks that involve interactions with the environment, such as planning and control.

\begin{figure}
\centering
\includegraphics[width=1.0\linewidth]{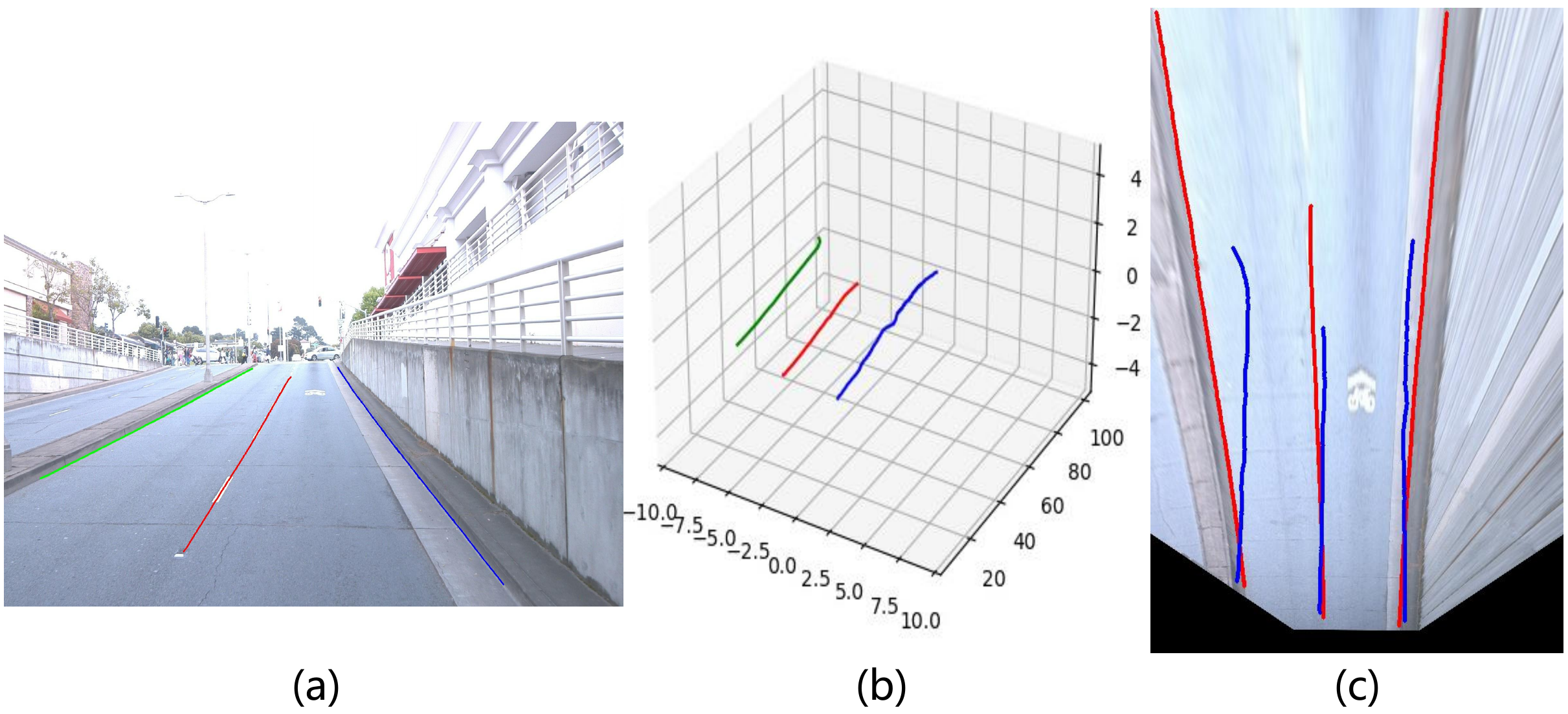}
\caption{The illustration of lanes lines in different views. (a) The lane lines in image view. (b) The lane lines in 3D space. (c) The lane lines in BEV. The red lane lines in (c) are obtained by projecting those in (a) with IPM. The blue lane lines in (c) are obtained by projecting those in (b) with height collapse.}
\label{Fig:example}
\end{figure}

One typical 3D lane detection pipeline is to first detect lane lines in image view, and then project them into BEV \cite{wang2014approach,meyer2018deep,neven2018towards,su2021structure,yan2022once}. The projection is usually achieved by inverse perspective mapping (IPM) which assumes a flat road surface. However, as shown in Figure \ref{Fig:example}, IPM will cause the projected lane lines to diverge or converge in case of uphill or downhill,  due to the neglect of the changes in road height. To solve this problem, SALAD \cite{yan2022once} predicts the real depth of lane lines along with their image-view positions, and then project them into 3D space with camera in/extrinsics. However, depth estimation is inaccurate at a distance, affecting the projection accuracy.

State-of-the-art methods tend to predict the 3D structure of lane lines directly from BEV \cite{garnett20193d,efrat20203d,guo2020gen,chen2022persformer,li2022reconstruct}. They first transform an image-view feature map into BEV by aid of IPM, and then detect lane lines based on the BEV feature map. However, as shown in Figure \ref{Fig:example}(c), due to the planar assumption of IPM, the ground-truth 3D lane lines (blue lines) are not aligned with the underlying BEV lane features (red lines), when encountering uneven roads. To solve this problem, some methods \cite{guo2020gen,chen2022persformer,li2022reconstruct} represent the ground-truth 3D lane lines in a virtual top-view by first projecting them onto the image plane, and then projecting them onto a flat road plane via IPM (red lines in Fig. \ref{Fig:example}(c)). Then these methods predict the real height of lane lines along with their positions in the virtual top-view, and finally project them into 3D space with a geometric transformation. However, the accuracy of the predicted heights significantly impacts the transformed BEV positions, which affects the model's robustness. Moreover, the separate view transformation and lane detection lead to cumulative error and increased complexity.

To overcome these limitations of current methods, we propose an efficient transformer for 3D lane detection. Unlike the vanilla transformer, our model incorporates a decomposed attention mechanism to learn lane and BEV representations simultaneously in a supervised manner. The mechanism decomposes the cross-attention between image-view and BEV features into the one between image-view and lane features, and the one between BEV and lane features. We supervise the decomposed cross-attentions with ground-truth lane lines, where we obtain the 2D and 3D lane predictions by applying the lane features to the image-view and BEV features, respectively.

To achieve this, we generate dynamic kernels for each lane line from the lane features. We then use these kernels to convolve the image-view and BEV feature maps, and obtain the image-view and BEV offset maps, respectively. The offset maps predict the offsets of each pixel to its nearest lane point in 2D and 3D space, which we then process with a voting algorithm to obtain the final 2D and 3D lane points, respectively. 
Since the view transformation is learned from data with a supervised cross-attention, it is more accurate than the IPM-based ones. Additionally, the lane and BEV features can dynamically adjust to each other with cross-attention, resulting in more accurate lane detection than the two separate stages. Our decomposed cross-attention is more efficient than the vanilla cross attention between image-view and BEV features. The experiments on two benchmark datasets, including OpenLane and ONCE-3DLanes demonstrate the effectiveness and efficiency of our method.

\section{Related Work}

\subsection{3D Lane Detection in Image View}
3D lane detection can be achieved in image view. Some methods first detect 2D lane lines in image view, and then project them into bird's-eye-view (BEV) \cite{wang2014approach,meyer2018deep,neven2018towards,su2021structure,yan2022once}. There are various methods proposed to tackle the problem of 2D lane detection, including the anchor-based methods \cite{zheng2022clrnet,tabelini2021keep,chen2019pointlanenet}, the parameter-based methods \cite{liu2021end,tabelini2021polylanenet} and the segmentation-based methods \cite{qu2021focus,ko2021key,liu2021condlanenet,li2021hdmapnet,neven2018towards}. As for lane line projection, some methods \cite{wang2014approach,meyer2018deep,neven2018towards,su2021structure} use inverse perspective mapping (IPM) which will cause the projected lane lines to diverge or converge when encountering uneven roads due to the planar assumption. To solve this problem, SALAD \cite{yan2022once} predicts the real depth of lane lines along with their image-view positions, and then project them into BEV with camera in/extrinsics. However, depth estimation is inaccurate at a distance, affecting the projection accuracy. Other methods predict the 3D structures of lane lines directly from image view. For example, CurveFormer \cite{bai2022curveformer} applies a transformer to predict the 3D curve parameters of lane lines directly from the image-view features. Anchor3DLane \cite{huang2023anchor3dlane} projects the lane anchors defined in 3D space onto the image-view feature map and extracts their features for classification and regression. However, these methods are limited by the low resolution of the image-view features in the distance.

\subsection{3D Lane Detection in Bird's-Eye-View}
Another way of 3D lane detection is to first transform an image view feature map into BEV, and then detect lane lines based on the BEV feature map \cite{garnett20193d,efrat20203d,guo2020gen,chen2022persformer,li2022reconstruct,wang2022bev}, where the view transformation is usually based on IPM. For example, some methods \cite{garnett20193d,efrat20203d,guo2020gen,li2022reconstruct}
adopt a spatial transformer network (STN) \cite{jaderberg2015spatial} for view transformation, where the sampling grid of STN is generated with IPM. PersFormer \cite{chen2022persformer} adopts a deformable transformer \cite{zhu2020deformable} for view transformation, where the reference points of the transformer decoder are generated with IPM. However, due to the planar assumption of IPM, the ground-truth 3D lane lines are not aligned with the underlying BEV lane features, when encountering uneven roads. To solve this problem, some methods \cite{guo2020gen, li2022reconstruct, chen2022persformer} represent the ground-truth 3D lane lines in a virtual top-view by first projecting them onto the image plane, and then projecting the result onto a flat ground with IPM. Then they predict the real height of lane lines along with their positions in the virtual top-view, and then project them into 3D space with a geometric transformation. However, the accuracy of the predicted heights significantly impacts the transformed BEV positions, affecting the model’s robustness. BEV-LaneDet \cite{wang2022bev} applies multi-layer perceptron (MLP) to achieve better view transformation. However, its parameter size is very large.

\subsection{Efficient Attention in Transformer}
The attention mechanism in transformer requires pairwise similarity calculation between queries and keys, which becomes complex for a large number of them. To solve this problem, some methods \cite{huang2019ccnet,zhu2020deformable,beltagy2020longformer,wang2021anchor,liu2021swin} focus on only a subset of the keys for each query instead of the whole set, when computing the attention matrix. For example, CCNet \cite{huang2019ccnet} proposes an attention module which harvests the contextual information of all pixels only on their criss-cross path. Deformable DETR \cite{zhu2020deformable} propose an attention module which only attends to a small set of key points sampled around the learned reference points. Swin Transformer \cite{liu2021swin} proposes a shifted windowing module which limits the self-attention to non-overlapping local windows while also allows for cross-window connection. Other methods \cite{wang2020linformer,chen2020compressed,xiong2021nystromformer} apply low-rank approximation to accelerate the computation of attention matrix. For example, LinFormer \cite{wang2020linformer} uses linear layers to project the original high resolution keys and values to a low resolution, which are attended by all queries. Nystr{\"o}mformer \cite{xiong2021nystromformer} adopts Nystr{\"o}m method \cite{baker1977numerical} to reconstruct the original attention matrix, thereby reducing the computation. 
While Nystr{\"o}mformer uses randomly sampled features for low-rank decomposition, our method decomposes the original attention matrix into two low-rank parts according to lane queries, and each part can be supervised with ground-truth, which is more suitable for the 3D lane detection task. Existing approximation of transformers usually sacrifices some accuracy, while our method has better performance than the original transformer.

\section{Method}
In this section, we present the proposed efficient transformer for end-to-end 3D lane detection. We first present the overall framework, and then describe each component in detail, including an efficient transformer module, a lane detection head and a bipartite matching loss.

\begin{figure*}
\centering
\includegraphics[width=1.0\linewidth]{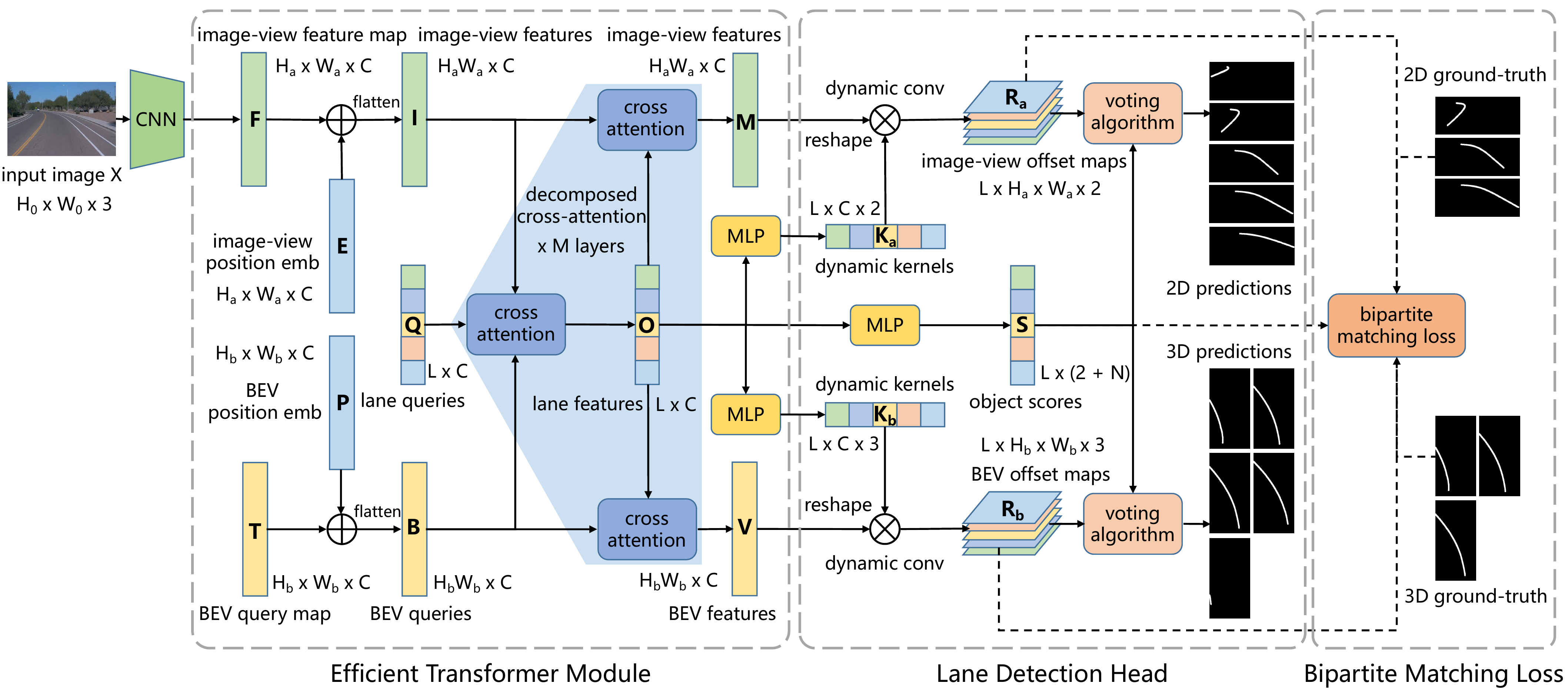}
\caption{Overview of the proposed 3D lane detection framework. The framework includes a CNN backbone to extract an image-view feature map from the input image, an efficient transformer module to generate lane and BEV features simultaneously from the image-view features, a lane detection head to detect the 2D and 3D lane lines from the generated lane, image-view and BEV features, and a bipartite matching loss for model training. The \textbf{dotted} arrow parts are only engaged in training.}
\label{Fig:framework}
\end{figure*}

\begin{figure}
\centering
\includegraphics[width=1.0\linewidth]{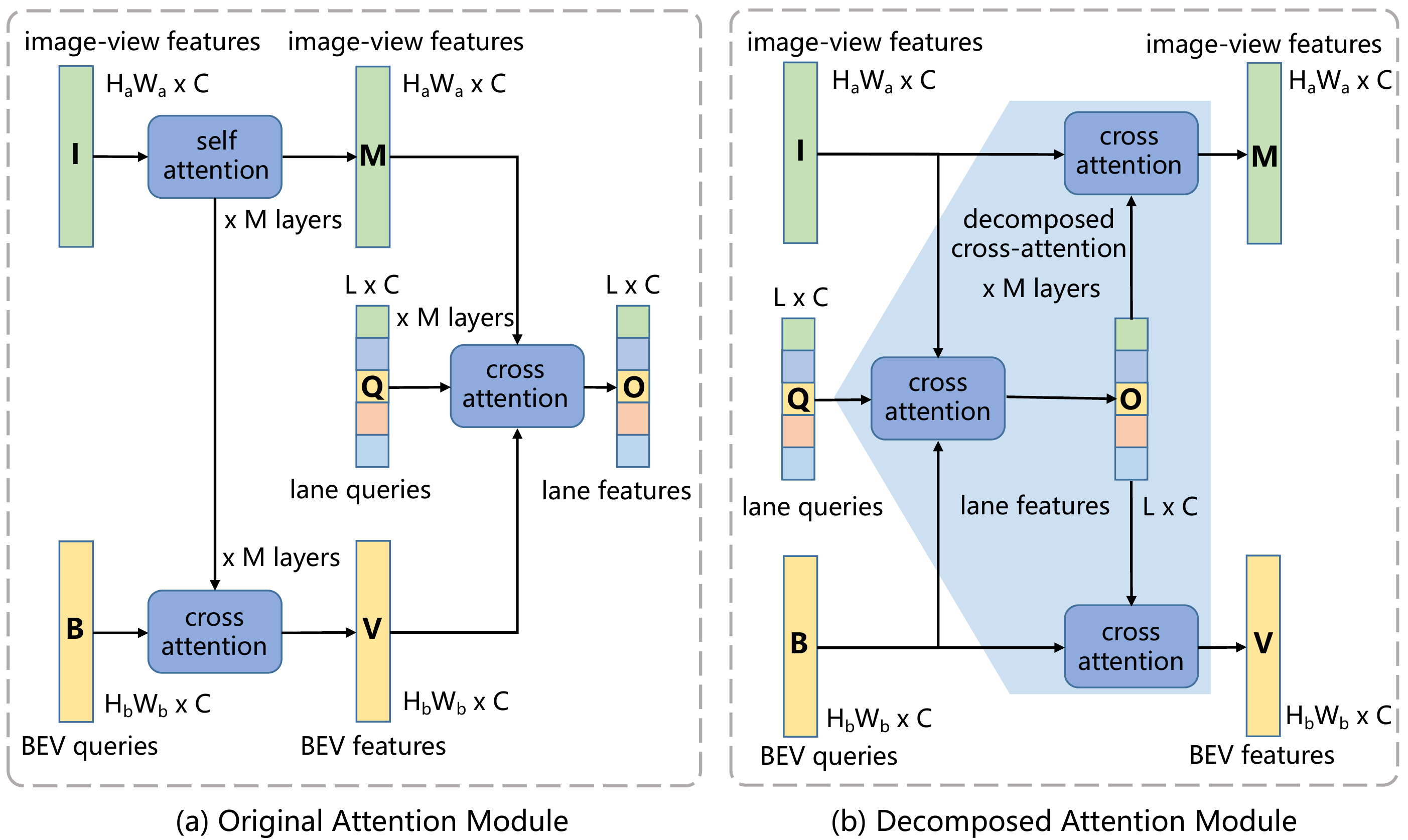}
\caption{The comparison of learning lane and BEV representations between the original attention module and the decomposed attention module.}
\label{Fig:frame_comp}
\end{figure}

\begin{figure}
\centering
\includegraphics[width=1.0\linewidth]{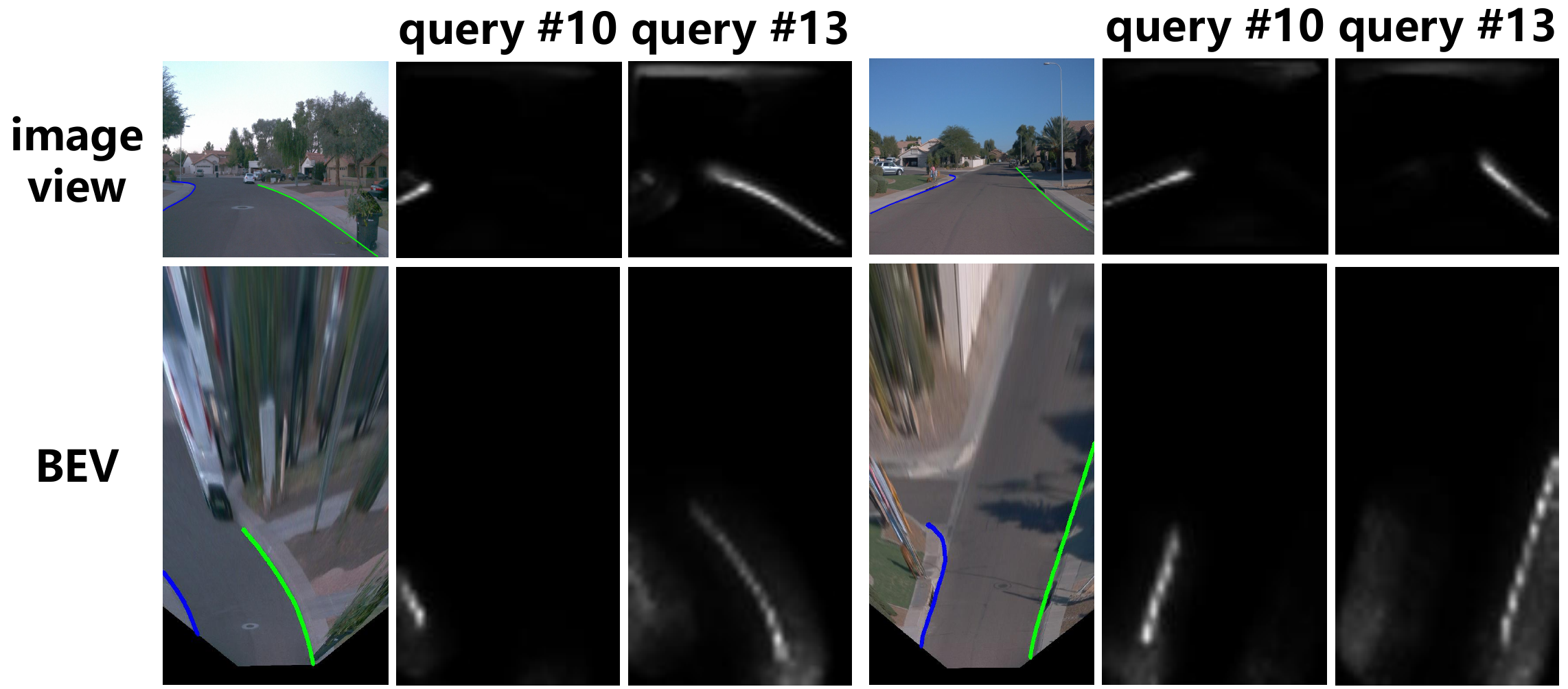}
\caption{Visualization of attention maps between lane features \textbf{O} and image-view features \textbf{M}, and between \textbf{O} and BEV features \textbf{V}.}
\label{Fig:attention}
\end{figure}

\subsection{Overall Framework}
The overall framework is shown in Figure \ref{Fig:framework}. It starts with a CNN backbone to extract an image-view feature map from the input image. Then the efficient transformer module learns the lane and bird's-eye view (BEV) features from the image-view features with a decomposed cross-attention mechanism. The image-view and BEV features are added with a position embeddings with their respective position encoders. Next, the lane detection head generates a set of dynamic kernels and object scores for each lane line using the lane features. These kernels are then used to convolve the image-view and BEV feature maps, producing image-view and BEV offset maps, respectively. The two sets of offset maps are processed with a voting algorithm to obtain the final 2D and 3D lane points, respectively. To train the model, a bipartite matching loss is calculated between the 2D/3D predictions and ground-truths. Thanks to the decomposed cross-attention mechanism, our framework achieves more accurate view transformation and lane detection while maintaining real-time efficiency.

\subsection{Efficient Transformer Module}
Here we describe how to learn the lane and BEV features simultaneously with the decomposed cross-attention mechanism. As shown in Figure \ref{Fig:framework}, given an input image $\textbf{X} \in \mathbb{R}^{H_0 \times W_0 \times 3}$, we first adopt a CNN backbone to extract an image-view feature map $\textbf{F} \in \mathbb{R}^{H_a \times W_a \times C}$, where $H_a$, $W_a$ and $C$ are the height, width, and channels of $\textbf{F}$, respectively. The feature map $\textbf{F}$ is added with a position embedding $\textbf{E} \in \mathbb{R}^{H_a \times W_a \times C}$ generated by a position encoder (described in Sec. \ref{Sec:embedding}), and then flattened into a sequence $\textbf{I} \in \mathbb{R}^{H_aW_a \times C}$. Then we initialize a BEV query map $\textbf{T} \in \mathbb{R}^{H_b \times W_b \times C}$ with learnable parameters, which is also added with a position embedding $\textbf{P} \in \mathbb{R}^{H_b \times W_b \times C}$ generated by another position encoder, and then flattened into a sequence $\textbf{B} \in \mathbb{R}^{H_bW_b \times C}$.

After obtaining the image-view features and the BEV queries, we initialize a set of lane queries $\textbf{Q} \in \mathbb{R}^{L \times C}$ with learnable parameters, representing $L$ different prototypes of lane lines. Then the lane features $\textbf{O} \in \mathbb{R}^{L \times C}$ are learned from the image-view features $\textbf{I}$ and BEV queries $\textbf{B}$ with a cross-attention. Let $\textbf{O}_i \in \mathbb{R}^C$ denote the $i$-th lane feature corresponds to the $i$-th lane query $\textbf{Q}_i$, $\textbf{O}_i$ can be obtained by
\begin{equation}
    \begin{aligned}
       \textbf{Q}_i + \sum_{j=1}^{H_aW_a} f_o(\textbf{Q}_i, \textbf{I}_j) g_o(\textbf{I}_j) + \sum_{k=1}^{H_bW_b} f_o(\textbf{Q}_i, \textbf{B}_k) g_o(\textbf{B}_k),
    \end{aligned}
    \label{Eq:1}
\end{equation}
where $\textbf{I}_j, \textbf{B}_k \in \mathbb{R}^C$ are the features at the $j$-th and $k$-th position of $\textbf{I}$ and $\textbf{B}$, respectively; $g_o(\textbf{I}_j) = \textbf{W}_g \textbf{I}_j$ is a linear transformation function where $\textbf{W}_g \in \mathbb{R}^{C \times C}$ is a learnable weight matrix; $f_o(\textbf{Q}_i, \textbf{I}_j)$ computes a pairwise similarity between $\textbf{Q}_i$ and $\textbf{I}_j$ as follows:
\begin{equation}
    \begin{aligned}
        f_o(\textbf{Q}_i, \textbf{I}_j) = \frac{exp(\textbf{Q}_{i}^\intercal \textbf{W}_{\theta}^\intercal \textbf{W}_{\phi} \textbf{I}_{j})}{\sum_{j'=1}^{H_aW_a} exp(\textbf{Q}_{i}^\intercal \textbf{W}_{\theta}^\intercal \textbf{W}_{\phi} \textbf{I}_{j'})},
    \end{aligned}
    \label{Eq:2}
\end{equation}
where $\textbf{W}_{\theta}, \textbf{W}_{\phi} \in \mathbb{R}^{C \times C}$ are learnable weight matrices.

Then the BEV features $\textbf{V}$ are constructed from the lane features $\textbf{O}$ with a cross-attention as follows:
\begin{equation}
    \begin{aligned}
        \textbf{V}_i = \textbf{B}_i + \sum_{j=1}^L f_v(\textbf{B}_i, \textbf{O}_j) g_v(\textbf{O}_j),
    \end{aligned}
    \label{Eq:3}
\end{equation}
where $g_v(\cdot)$ and $f_v(\cdot, \cdot)$ have the same form as $g_o(\cdot)$ and $f_o(\cdot, \cdot)$ in Eq. (\ref{Eq:1}), respectively, except for their learnable weight matrices. In this way, the original cross-attention between the BEV features $\textbf{V}$ and the image-view features $\textbf{I}$ shown in Figure \ref{Fig:frame_comp} is decomposed into the one between the image-view features $\textbf{I}$ and the lane features $\textbf{O}$, and the one between the lane features $\textbf{O}$ and the BEV features $\textbf{V}$. 

Compared to the original one, the decomposed cross-attention provides three benefits. Firstly, it enables better view transformation by supervising the decomposed cross-attentions with 2D and 3D ground-truth lane lines. The 2D and 3D lane predictions are obtained by applying the lane features $\textbf{O}$ to the image-view features $\textbf{I}$ and the BEV features $\textbf{V}$, respectively. Secondly, it improves the accuracy of 3D lane detection by enabling the dynamic adjustment between the lane features $\textbf{O}$ and the BEV features $\textbf{V}$ through cross-attention. Thirdly, it significantly reduces the computation amount and enables real-time efficiency.

Similarly, the image-view features $\textbf{I}$ are updated with the object features $\textbf{O}$ with a cross-attention as follows:
\begin{equation}
    \begin{aligned}
        \textbf{M}_i = \textbf{I}_i + \sum_{j=1}^L f_m(\textbf{I}_i, \textbf{O}_j) g_m(\textbf{O}_j),
    \end{aligned}
    \label{Eq:4}
\end{equation}
where $\textbf{M}$ is the updated image-view features, $g_m(\cdot)$ and $f_m(\cdot, \cdot)$ has the same form as $g_o(\cdot)$ and $f_o(\cdot, \cdot)$ in Eq. (\ref{Eq:1}), respectively, except for their learnable weight matrices. In this way, the original self-attention within the image-view features $\textbf{I}$ shown in Figure \ref{Fig:frame_comp} is decomposed into the back-and-forth cross-attention between the image-view features $\textbf{I}$ and the lane features $\textbf{O}$. 

Compared to the original one, the decomposed self-attention improves the accuracy of 2D lane detection by enabling the dynamic adjustment between the lane features $\textbf{O}$ and the image-view features $\textbf{I}$ through cross-attention. Moreover, since the image-view features $\textbf{M}$ and the BEV features $\textbf{V}$ are both constructed from the object features $\textbf{O}$, they can better align with each other.

\subsection{Position Embedding}
\label{Sec:embedding}
Here we describe how to generate the position embeddings for transfomer inputs. For the image-view position embedding $\textbf{E} \in \mathbb{R}^{H_a \times W_a \times C}$, we first constructs a 3D coordinate grid $\textbf{G} \in \mathbb{R}^{H_a \times W_a \times D \times 4}$ in camera space, where $D$ is the number of discretized depth bins. Each point in $\textbf{G}$ can be represented as $p_j = (u_j \times d_j, v_j \times d_j, d_j, 1)^\intercal$, where $(u_j, v_j)$ is the corresponding pixel coordinate in the image-view feature map $\textbf{F}$, $d_j$ is the corresponding depth value. 

Then the grid $\textbf{G}$ is transformed into a grid $\textbf{G}' \in \mathbb{R}^{H_a \times W_a \times D \times 4}$ in 3D space as follows:
\begin{equation}
    \begin{aligned}
        p_j' = K^{-1}p_j,
    \end{aligned}
    \label{Eq:5}
\end{equation}
where $p_j'$ is the point in $\textbf{G}'$ corresponding to the point $p_j$ in $\textbf{G}$, $K \in \mathbb{R}^{4 \times 4}$ is a projection matrix from the 3D space to the camera space. Then we predict the depth distributions $\textbf{D} \in \mathbb{R}^{H_a \times W_a \times D}$ from $\textbf{F}$, indicating the probability of each pixel belonging to each depth bin. Then the position embedding $\textbf{E}$ is obtained as follows:
\begin{equation}
    \begin{aligned}
        \textbf{E}_{uv} = \left[ \sum_{d=1}^D \textbf{D}_{uvd} (\textbf{G}'_{uvd} \textbf{W}_1 + \textbf{b}_1) \right] \textbf{W}_2 + \textbf{b}_2,
    \end{aligned}
    \label{Eq:6}
\end{equation}
where $\textbf{E}_{uv} \in \mathbb{R}^C$ is the pixel embedding at $u$-th row and $v$-th column of $\textbf{F}$; $\textbf{W}_1 \in \mathbb{R}^{4 \times C / 4}$, $\textbf{b}_1 \in \mathbb{R}^{C/4}$, $\textbf{W}_2 \in \mathbb{R}^{C / 4 \times C}$ and $\textbf{b}_2 \in \mathbb{R}^C$ are learnable weights.

For the BEV position embedding $\textbf{P} \in \mathbb{R}^{H_b \times W_b \times C}$, we first constructs a 3D coordinate grid $\textbf{H} \in \mathbb{R}^{H_b \times W_b \times Z \times 4}$ in 3D space, where $Z$ is the number of discretized height bins. Each point in $\textbf{H}$ can be represented as $p_i = (x_i \times s_x + r_x^0, y_i \times s_y + r_y^0, z_i, 1)^\intercal$, where $(x_i, y_i)$ is the corresponding pixel coordinate in the BEV query map $\textbf{T}$; $z_i$ is the corresponding height value; $s_x = (r_x^1 - r_x^0) / W_b$ and $(r_x^0, r_x^1)$ are the scale factor and range for the $x$ direction in 3D space, respectively (similar for $s_y$ and $r_y$). 

Then we predict the height distributions $\textbf{Z} \in \mathbb{R}^{H_b \times W_b \times Z}$ from $\textbf{T}$, indicating the probability of each pixel belonging to each height bin. Then the position embedding $\textbf{P}$ is obtained as follows:
\begin{equation}
    \begin{aligned}
        \textbf{P}_{xy} = \left[ \sum_{z=1}^Z \textbf{Z}_{xyz} (\textbf{H}_{xyz} \textbf{W}_1 + \textbf{b}_1) \right] \textbf{W}_2 + \textbf{b}_2,
    \end{aligned}
    \label{Eq:7}
\end{equation}
where $\textbf{P}_{xy} \in \mathbb{R}^C$ is the pixel embedding at $x$-th row and $y$-th column of $\textbf{T}$; $\textbf{W}_1$, $\textbf{b}_1$, $\textbf{W}_2$ and $\textbf{b}_2$ are the same as those in Eq. (\ref{Eq:6}).

\begin{algorithm}[t!]
\caption{The voting algorithm.}
\label{Alg:voting}
\SetKwData{Or}{\textbf{or}}
\DontPrintSemicolon
\KwIn {$\textbf{R}_b$: BEV offset maps, $\textbf{S}$: object scores, \quad \quad $r_x$, $r_y$: $x$ and $y$ ranges in 3D space, \quad \quad \quad \quad $t$: object threshold, $w$: lane width threshold}
Initialize a vote count matrix $\textbf{U}_b \in \mathbb{R}^{L \times H_b \times W_b}$\;
Initialize a BEV lane line set $\textbf{Y}_b = \emptyset$\;
\For{$i=0$ \KwTo $L-1$}{
    \For{$j=0$ \KwTo $H_b-1$}{
      \For{$k=0$ \KwTo $W_b-1$}{
        $x, y \gets \left[k + \textbf{R}_b^{ijk0} \right], \left[j + \textbf{R}_b^{ijk1} \right]$\;
        $s \gets exp(-(\frac{(\textbf{R}_b^{ijk0})^2 + (\textbf{R}_b^{ijk1})^2)}{2w^2})$\;
        $\textbf{U}_b^{iyx} \gets \textbf{U}_b^{iyx} + s$\;
      }
    }
    Initialize a BEV lane point set $\textbf{L}_b = \emptyset$\;
    \For{$j=0$ \KwTo $H_b-1$}{
      \For{$k=0$ \KwTo $W_b-1$}{
        \If{$\textbf{U}_b^{ijk} \geq w$}{
          $x \gets k \times (r_x^1 - r_x^0) / W_b + r_x^0$\;
          $y \gets j \times (r_y^1 - r_y^0) / H_b + r_y^0$\;
          $z \gets \textbf{R}_b^{ijk2}$\;
          $\textbf{L}_b \gets \textbf{L}_b \cup \{(x, y, z)\}$\;
        }
      }
    }
    \If{$\textbf{S}_{i1} \geq t$}{
      $\textbf{Y}_b \gets \textbf{Y}_b \cup \{\textbf{L}_b\}$\;
    }
}
\KwOut{the BEV lane line set $\textbf{Y}_b$.}
\end{algorithm}

\subsection{Lane Detection Head}
Here we describe the lane detection head. We first apply two multi-layer perceptrons (MLPs) to the lane features $\textbf{O}$ to generate two sets of dynamic kernels $\textbf{K}_a \in \mathbb{R}^{L \times C \times 2}$ and $\textbf{K}_b \in \mathbb{R}^{L \times C \times 3}$, respectively. Then we apply $\textbf{K}_a$ and $\textbf{K}_b$ to convolve the image-view features $\textbf{M}$ and the BEV features $\textbf{V}$, to obtain the image-view offset map $\textbf{R}_a \in \mathbb{R}^{L \times H_a \times W_a \times 2}$ and the BEV offset map $\textbf{R}_b \in \mathbb{R}^{L \times H_b \times W_b \times 3}$, respectively. $\textbf{R}_a$ predicts the horizontal and vertical offsets of each pixel to its nearest lane point in image view; $\textbf{R}_b$ predicts the offsets in $x$ and $y$ directions of each pixel to its nearest lane point in BEV, along with the real height of the lane point. 

Then we apply another MLP to the lane features $\textbf{O}$ to generate the object scores $\textbf{S} \in \mathbb{R}^{L \times (2 + N)}$, including the probabilities of background, foreground, and $N$ lane classes. Then the image-view offset maps $\textbf{R}_a$ and the BEV offset maps $\textbf{R}_b$ are processed with a voting algorithm to obtain the 2D and 3D lane points, respectively. The process of $\textbf{R}_b$ is shown in Algorithm \ref{Alg:voting} (similar for $\textbf{R}_a$, except for the removal of $z$ and $r$). The voting algorithm casts votes for the predicted lane points of all pixels, and then selects those with votes exceeding a lane width threshold $w$ to form a predicted lane line. Finally, only the predicted lane lines with a foreground probability exceeding an object threshold $t$ are retained as output.

\subsection{Bipartite Matching Loss}
Here we introduce the design of loss functions. First, we need to compute a pair-wise matching cost $L_{match}(\textbf{Y}_i, \textbf{Y}_j^*)$ between $L$ predicted lane lines $\textbf{Y} = \{\textbf{Y}_i\}_{i=1}^L$ and $M$ ground-truth lane lines $\textbf{Y}^* = \{\textbf{Y}_i^*\}_{i=1}^L$, where each $\textbf{Y}_i$ contains object scores $\textbf{S}$, image-view offset maps $\textbf{R}_a$ and BEV offset maps $\textbf{R}_b$ (same for $\textbf{Y}_i^*$). The matching cost includes an object cost, a classification cost and an offset map cost. The object cost is defined as follows:
\begin{equation}
    L_{obj}(\textbf{Y}_i, \textbf{Y}_j^*) = - log(\textbf{S}_{i1}),
\end{equation}
where $\textbf{S}_{i1}$ is the predicted foreground probability of the $i$-th predicted lane line. The classification cost is defined as follows:
\begin{equation}
    L_{cls}(\textbf{Y}_i, \textbf{Y}_j^*) = - \sum_{k=0}^{N-1} \textbf{S}_{j(k+2)}^* \cdot log(\textbf{S}_{i(k+2)}),
\end{equation}
where $\textbf{S}_{i(k+2)}$ is the predicted probability of the $k$-th class of the $i$-th prediction, $\textbf{S}_{j(k+2)}^*$ is the corresponding label of the $j$-th ground-truth, $N$ is the number of classes. The offset map cost is defined as follows:
\begin{equation}
    \begin{aligned}
        L_{off}(\textbf{Y}_i, \textbf{Y}_j^*) = & \frac{1}{H_aW_a} \sum_{m=0}^{H_a-1}\sum_{n=0}^{W_a-1} \Vert \textbf{R}_a^{imn} - (\textbf{R}_a^{jmn})^*\Vert_1 \\ + & \frac{1}{H_bW_b} \sum_{m=0}^{H_b-1}\sum_{n=0}^{W_b-1} \Vert \textbf{R}_b^{imn} - (\textbf{R}_b^{jmn})^*\Vert_1,
    \end{aligned}
\end{equation}
where $\textbf{R}_a^i \in \mathbb{R}^{H_a \times W_a \times 2}$ and $\textbf{R}_b^i \in \mathbb{R}^{H_b \times W_b \times 3}$ are the predicted image-view and BEV offset maps of the $i$-th prediction, $(\textbf{R}_b^j)^*$ and $(\textbf{R}_a^j)^*$ are the corresponding labels of the $j$-th ground-truth. The final pair-wise matching cost is defined as follows:
\begin{equation}
    \begin{aligned}
        L_{match}(\textbf{Y}_i, \textbf{Y}_j^*) & = \lambda_{obj} L_{obj}(\textbf{Y}_i, \textbf{Y}_j^*) + \lambda_{cls} L_{cls}(\textbf{Y}_i, \textbf{Y}_j^*) \\ & +  \lambda_{off} L_{off}(\textbf{Y}_i, \textbf{Y}_j^*),
    \end{aligned}
\end{equation}
where $\lambda_{obj}$, $\lambda_{cls}$ and $\lambda_{off}$ are the balance weights for $L_{obj}$, $L_{cls}$ and $L_{off}$, respectively. Then we find an optimal injective function $z: \{\textbf{Y}_i\}_{i=1}^L \rightarrow \{\textbf{Y}_j^*\}_{j=1}^M$, where $z(j)$ is the index of the prediction assigned to the $j$-th ground-truth, by minimizing the matching cost as follows:
\begin{equation}
    \underset{z}{{\arg\min} \, \hat{z}} = \sum_{j=1}^N L_{match}(\textbf{Y}_{z(j)}, \textbf{Y}_j^*).
\end{equation}
This objective function can be solved by the Hungarian algorithm \cite{kuhn1955hungarian}. After obtaining the injective mapping $z$, we can compute the final loss function as follows:
\begin{equation}
    \begin{aligned}
        loss & = \frac{1}{M} \sum_{j=1}^{M} \big[\lambda_{obj} L_{obj}(\textbf{Y}_{z(j)}, \textbf{Y}_j^*) +  \lambda_{cls} L_{cls}(\textbf{Y}_{z(j)}, \textbf{Y}_j^*) \\ + & \lambda_{off} L_{off}(\textbf{Y}_{z(j)}, \textbf{Y}_j^*) \big] + \frac{1}{L-M} \sum_{i \notin \textbf{C}} \lambda_{obj} L_{obj}(\textbf{Y}_i, \phi),
    \end{aligned}
\end{equation}
where $\textbf{C} = \{z(j)\}_{j=1}^M$ is the index set of the predictions matched to the ground-truths, $L_{obj}(\textbf{Y}_i, \phi) = -log(\textbf{S}_{i0})$, where $\textbf{S}_{i0}$ is the predicted background probability of the $i$-th prediction.

\section{Experiment}

\begin{table*}[t!]
	\centering 
	\resizebox{1.0\hsize}{!}{
    	\begin{tabular}{l l c c c c c c c}
    		\hline
    		\textbf{Method} & \textbf{Backbone} & \textbf{F1(\%)} & \textbf{Cate Acc.(\%)} & \textbf{X err. near (m)} & \textbf{X err. far (m)} & \textbf{Z err. near (m)} & \textbf{Z err. far (m)} \\
    		\hline
    		3D-LaneNet \cite{garnett20193d}            & VGG-16          & 44.1      & -          & 0.479      & 0.572       & 0.367      & 0.443        \\
    		GenLaneNet \cite{guo2020gen}               & ERFNet          & 32.3      & -          & 0.591      & 0.684       & 0.411      & 0.521        \\
    		PersFormer \cite{chen2022persformer}       & EfficientNet    & 50.5      & \bf{92.3}  & 0.485      & 0.553       & 0.364      & 0.431        \\
            CurveFormer \cite{bai2022curveformer}      & EfficientNet    & 50.5      & -          & 0.340      & 0.772       & 0.207      & 0.651        \\
            Anchor3DLane \cite{huang2023anchor3dlane}  & ResNet-18       & 54.3      & 90.7       & 0.275      & 0.310       & 0.105      & 0.135        \\
            BEV-LaneDet \cite{wang2022bev}             & ResNet-18       & 57.8      & -          & 0.318      & 0.705       & 0.245      & 0.629        \\
            BEV-LaneDet \cite{wang2022bev}             & ResNet-34       & 58.4      & -          & 0.309      & 0.659       & 0.244      & 0.631        \\
    		\hline
            \textbf{Ours}                              & ResNet-18       & 60.7      & 89.2       & 0.287      & 0.358       & 0.112      & 0.143        \\
            \textbf{Ours}                              & ResNet-34       & 62.8      & 91.2       & 0.254      & 0.322       & 0.105      & 0.135        \\
            \textbf{Ours}                              & EfficientNet    & \bf{63.8} & 91.5       & \bf{0.245} & \bf{0.304}  & \bf{0.104} & \bf{0.129}   \\
    		\hline
    	\end{tabular}
 	}
	\caption{Comparison with state-of-the-art methods on OpenLane validation set. We report the F-score, category accuracy, and regression errors in x and z directions of the models.}
	\label{Tab1}
	\vspace{-1mm}
\end{table*}

\begin{table*}[t!]
	\centering 
	\resizebox{1.0\hsize}{!}{
    	\begin{tabular}{l l c c c c c c c}
    		\hline
    		\textbf{Method} & \textbf{Backbone} & \textbf{All} & \textbf{Up \& Down} & \textbf{Curve} & \textbf{Extreme Weather} & \textbf{Night} & \textbf{Intersection} & \textbf{Merge \& Split} \\
    		\hline
    		3D-LaneNet \cite{garnett20193d}            & VGG-16          & 44.1      & 40.8       & 46.5       & 47.5        & 41.5       & 32.1         & 41.7             \\
    		GenLaneNet \cite{guo2020gen}               & ERFNet          & 32.3      & 25.4       & 33.5       & 28.1        & 18.7       & 21.4         & 31.0             \\
    		PersFormer \cite{chen2022persformer}       & EfficientNet    & 50.5      & 42.4       & 55.6       & 48.6        & 46.6       & 40.0         & 50.7             \\
            CurveFormer \cite{bai2022curveformer}      & EfficientNet    & 50.5      & 45.2       & 56.6       & 49.7        & 49.1       & 42.9         & 45.4             \\
            Anchor3DLane \cite{huang2023anchor3dlane}  & ResNet-18       & 54.3      & 47.2       & 58.0       & 52.7        & 48.7       & 45.8         & 51.7             \\
            BEV-LaneDet \cite{wang2022bev}             & ResNet-34       & 58.4      & 48.7       & 63.1       & 53.4        & 53.4       & 50.3         & 53.7             \\
    		\hline
            \textbf{Ours}                              & ResNet-18       & 60.7      & 56.9       & 69.4       & 53.8        & 55.3       & 53.8         & 60.1             \\
            \textbf{Ours}                              & ResNet-34       & 62.8      & 56.9       & 71.0       & 54.3        & 57.9       & 55.9         & 63.1             \\
            \textbf{Ours}                              & EfficientNet    & \bf{63.8} & \bf{57.6}  & \bf{73.2}  & \bf{57.3}   & \bf{59.7}  & \bf{57.0}    & \bf{64.9}        \\
    		\hline
    	\end{tabular}
 	}
	\caption{Comparison with state-of-the-art methods on OpenLane validation set. We report the F-score for the whole validation set and under different scenarios, including curve, intersection, night, extreme weather, merge and split, and up and down.}
	\label{Tab2}
	\vspace{-3mm}
\end{table*}

\begin{table}[t!]
	\centering 
	\resizebox{1.0\hsize}{!}{
    	\begin{tabular}{l l c c c c}
    		\hline
    		\textbf{Method} & \textbf{Backbone} & \textbf{F1(\%)} & \textbf{Prec.(\%)} & \textbf{Rec.(\%)} & \textbf{CD Err.(m)} \\
    		\hline
    		3D-LaneNet \cite{garnett20193d}            & VGG-16          & 44.73      & 61.46      & 35.16      & 0.127                    \\
    		GenLaneNet \cite{guo2020gen}               & ERFNet          & 45.59      & 63.95      & 35.42      & 0.121                    \\
            SALAD \cite{yan2022once}                   & Segformer       & 64.07      & 75.90      & 55.42      & 0.098                    \\
    		PersFormer \cite{chen2022persformer}       & EfficientNet    & 74.33      & 80.30      & 69.18      & 0.074                    \\
            Anchor3DLane \cite{huang2023anchor3dlane}  & ResNet-18       & 74.87      & 80.85      & 69.71      & 0.060                    \\
    		\hline
            \textbf{Ours}                              & ResNet-18       & 79.67      & 82.66      & 76.89      & 0.057                    \\
            \textbf{Ours}                              & ResNet-34       & 80.49      & \bf{84.67} & 76.70      & 0.057                    \\
            \textbf{Ours}                              & EfficientNet    & \bf{80.84} & 84.50      & \bf{77.48} & \bf{0.056}               \\
    		\hline
    	\end{tabular}
 	}
	\caption{Comparison with state-of-the-art methods on ONCE-3DLanes test set. We report the F-score, precision, recall and chamfer distance (CD) error of the models.}
	\label{Tab3}
	\vspace{-3mm}
\end{table}

\subsection{Datasets}
We conduct experiments on two 3D lane detection benchmarks: OpenLane  \cite{chen2022persformer} and ONCE-3DLanes \cite{yan2022once}. OpenLane contains $160K$ and $40K$ images for training and validation sets, respectively. The validation set consists of six different scenarios, including curve, intersection, night, extreme weather, merge and split, and up and down. It annotates $14$ lane categories, including road edges, double yellow solid lanes, and so on. ONCE-3DLanes contains $200$K, $3$K, and $8$K images for training, validation, and testing, respectively, covering different time periods including morning, noon, afternoon and night, various weather conditions including sunny, cloudy and rainy days, as well as a variety of regions including downtown, suburbs, highway, bridges and tunnels.

\subsection{Evaluation Metrics}
Following \cite{chen2022persformer}, we adopt the F-score for regression and the accuracy of matched lanes for classification. We match predictions to ground truth using edit distance, where a predicted lane is only considered a true positive if 75\% of its y-positions have a point-wise distance less than the maximum allowed distance of $1.5$ meters. For ONCE-3DLanes, following \cite{yan2022once}, we adopt a two-stage evaluation metric to measure the similarity between predicted and ground-truth lanes. We first use the traditional IoU method to match lanes in the top-view. If the IoU is greater than a threshold ($0.3$), we use a unilateral Chamfer Distance to calculate the curves matching error. If the chamfer distance is less than a threshold ($0.3$m), we consider the predicted lane a true positive. At last, we use precision, recall, and F-score as metrics.

\subsection{Implementation Details}
We adopt ResNet-18, ResNet-34 \cite{he2016deep}, EfficientNet(-B7) \cite{tan2019efficientnet} with the pretrained weights from ImageNet \cite{deng2009imagenet} as the CNN backbones. The input images are augmented with random horizontal flipping and random rotation, and resized to $368 \times 480$. The spatial resolution of the BEV feature map is $50 \times 32$, representing a BEV space with the range of $\left[ -10, 10\right] \times \left[3, 103\right]$ meters along $x$ and $y$ directions, respectively. The BEV offset map is resized to $400 \times 256$ for final prediction. Optimization is done by AdamW \cite{loshchilov2017decoupled} with betas of $0.9$ and $0.999$, and weight decay of $1e^{-4}$. The batch size is set to $16$. We train the model for $50$ epochs. The base learning rate is initialized at $1e^{-4}$ and decayed to $1e^{-5}$ after $40$ epochs. The learning rate for the backbone is set to $0.1$ times of the base learning rate. The lane query number $L$ is set to $80$. The weights $\lambda_{obj}$, $\lambda_{cls}$, and $\lambda_{off}$ are set to $5$, $5$, and $1$, respectively, to balance different losses, i.e., bring them to the same scale. The object threshold $t$ is set to $0.7$. The lane width threshold $w$ is set to $16$.

\subsection{Results}

\begin{figure*}[t!]
	\centering
	\includegraphics[width=1.0\linewidth]{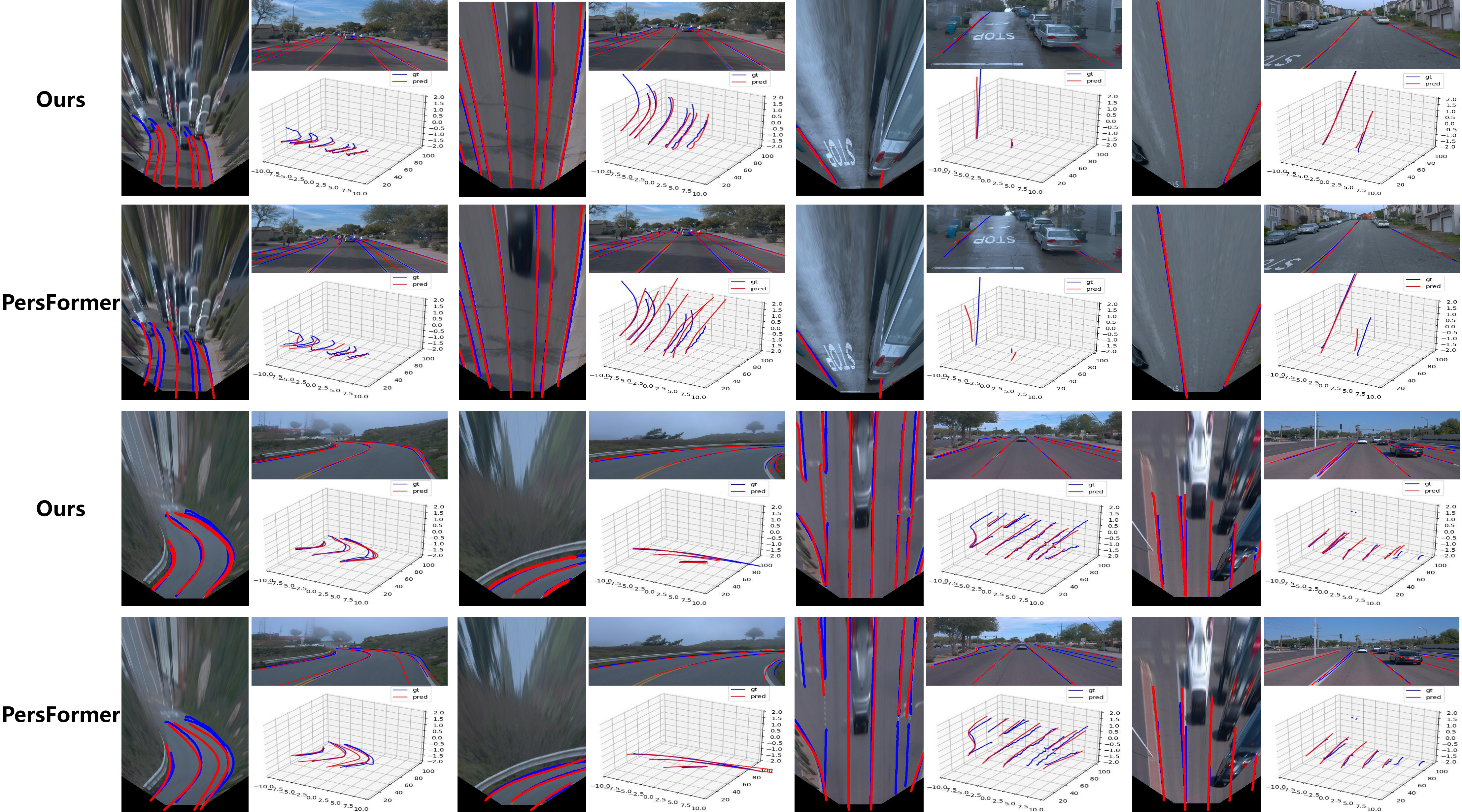}
	\caption{Visualization results on OpenLane. We choose PersFormer \cite{chen2022persformer} as our comparison method. We visualize the results under different scenarios, where the first two rows include uphill and downhill cases, and the last two rows include curved and forked lane lines. The blue lane lines are ground-truths and the red lane lines are predictions.}
	\label{Fig:visulization}
\end{figure*}

\paragraph{Performance on OpenLane Dataset.} The comparison results on OpenLane are shown in Table \ref{Tab1}. Using ResNet-18 as backbone, our method achieves a F-score of $60.7$, which is $10.2$, $6.4$ and $2.9$ points higher than that of PersFormer, Anchor3DLane and BEV-LaneDet, respectively. Our method also achieves the lowest prediction errors along the x, y and z directions. As shown in Table \ref{Tab2}, our method achieves the best performance in all of the six scenarios, demonstrating the robustness of our method. For example, for the ``Up \& Down", ``Curve", ``Intersection" and ``Merge \& Split" scenarios, using ResNet-34 as backbone, the F-scores of our method are $8.2$, $7.9$, $5.6$ and $9.4$ points higher than those of BEV-LaneDet, respectively. In Figure \ref{Fig:visulization}, we show qualitative comparison results on OpenLane, including upill, downhill, curved and forked scenarios. The comparison results demonstrate that our method can cope with the lane lines on uneven roads and those with complex topologies very well. This is because simultaneously learning lane and BEV features under supervision allows them to adjust with each other to generate more accurate results than the separate two-stage pipeline.

\paragraph{Performance on ONCE-3DLanes Dataset.} The comparison results on ONCE-3DLanes is shown in Table \ref{Tab3}. Using ResNet-18 as backbone, our method achieves a F-score of $79.67$, which is $15.60$, $5.34$ and $4.80$ points higher than that of SALAD, PersFormer and Anchor3DLane, respectively. Our method also achieves the lowest CD errors, demonstrating the good accuracy of the proposed method.

\subsection{Ablation Studies}

\begin{table*}[t!]
	\centering 
	\resizebox{1.0\hsize}{!}{
    	\begin{tabular}{c | c c c| c c c | c c c | c}
    		\hline
    		\multirow{2}*{\textbf{Exp.}} & \multirow{2}*{\textbf{\begin{tabular}[c]{@{}c@{}}IPM-based\\Attention\end{tabular}}} & \multirow{2}*{\textbf{\begin{tabular}[c]{@{}c@{}}Original\\Attention\end{tabular}}} & \multirow{2}*{\textbf{\begin{tabular}[c]{@{}c@{}}Decomposed\\Attention\end{tabular}}} & \multicolumn{3}{c|}{OpenLane} & \multicolumn{3}{c|}{ONCE-3DLanes} & \multirow{2}*{\textbf{FPS}} \\
            \cline{5-10} &           &           &           & \textbf{F1(\%)} & \textbf{Prec.(\%)} & \textbf{Rec.(\%)}  & \textbf{F1(\%)} & \textbf{Prec.(\%)} & \textbf{Rec.(\%)} &  \\
            \hline
            1            & \ding{52} & \ding{56} & \ding{56} & 58.6            & 63.6               & 54.3               & 77.5            & 81.3               & 74.1              & 25 \\
            2            & \ding{56} & \ding{52} & \ding{56} & 59.5            & 63.4               & 56.2               & 78.2            & 81.5               & 75.2              & 17 \\
            3            & \ding{56} & \ding{56} & \ding{52} & \bf{62.8}       & \bf{67.8}          & \bf{58.5}          & \bf{80.5}       & \bf{84.7}          & \bf{76.7}         & 45 \\
    		\hline
    	\end{tabular}
 	}
    \caption{The comparison results between different attention mechanisms on OpenLane and ONCE-3DLanes datasets. The backbone we use is ResNet-34.}
	\label{Tab4}
	\vspace{-3mm}
\end{table*}

\begin{table}[h]
	\centering 
	\resizebox{1.0\hsize}{!}{
    	\begin{tabular}{c | r r r | r r r}
    		\hline
    		\multirow{2}*{\textbf{\begin{tabular}[c]{@{}c@{}}Num of\\lane queries\end{tabular}}}  & \multicolumn{3}{c|}{OpenLane} & \multicolumn{3}{c}{ONCE-3DLanes} \\
            \cline{2-7}                             & \textbf{F1} & \textbf{Prec.} & \textbf{Rec.}  & \textbf{F1} & \textbf{Prec.} & \textbf{Rec.} \\
    		\hline
                20                                  & 59.4        & 64.4           & 55.2             & 78.1        & 82.5           & 74.2           \\
                40                                  & 61.3        & 66.2           & 57.1             & 79.4        & 83.8           & 75.4           \\
                80                                  & 62.8        & 67.8           & 58.5             & 80.5        & 84.7           & 76.7           \\
                160                                 & 63.7        & 68.7           & 59.4             & 81.1        & 85.4           & 77.2           \\
    		\hline
    	\end{tabular}
 	}
	\caption{Comparison of the results with different number of lane queries on OpenLane and ONCE-3DLanes datasets. The backbone we use is ResNet-34.}
	\label{Tab5}
	\vspace{-3mm}
\end{table}

\paragraph{Influence of decomposed attention.} We compare the results of the proposed decomposed attention (Fig. \ref{Fig:frame_comp} (b)), the original attention (Fig. \ref{Fig:frame_comp} (a)) and IPM-based attention. The IPM-based attention is used in PersFormer \cite{chen2022persformer} for view transformation, where the reference points of the deformable transformer are computed with IPM. As shown in Table \ref{Tab4}, the performance of the original attention is slightly better than that of the IPM-based attention, and the F1 scores of the decomposed attention are $3.3$ and $2.3$ points higher than those of the original attention on OpenLane and ONCE-3DLanes, respectively. This is because the decomposed attention allows for more accurate view transformation by supervising the cross-attention between the image-view and lane features, as well as the cross-attention between the lane and BEV features, using 2D and 3D ground-truths respectively. Additionally, it improves the accuracy of 3D lane detection by enabling dynamic adjustments between lane and BEV features through cross-attention. The speed of the decomposed attention is $2.6$ times that of the original attention, demonstrating the efficiency of the proposed decomposed attention mechanism.



\paragraph{Number of lane queries.} Here we investigate the influence of different number of lane queries. As shown in Table \ref{Tab5}, when we increase the number of lane queries from $20$ to $80$, the performance is improved. This demonstrates that increasing the number of lane queries is beneficial to the transformation between image-view and BEV features. When we further increase the number of lane queries from $80$ to $160$, the performance only increases slightly, which is because an excessive number of lane queries can cause redundancy.

\section{Conclusion}
In this paper, we propose an efficient transformer for 3D lane detection, utilizing a decomposed attention mechanism to simultaneously learn lane and BEV representations. The mechanism decomposes the cross-attention between image-view and BEV features into the one between image-view and lane features, and the one between lane and BEV features, both of which are supervised with ground-truth lane lines. This allows for a more accurate view transformation than IPM-based methods, and a more accurate lane detection than the traditional two-stage pipeline.

{\small
\bibliographystyle{ieee_fullname}
\bibliography{egbib}
}

\end{document}


\title{Supplementary Material}

\author{First Author\\
Institution1\\
Institution1 address\\
{\tt\small firstauthor@i1.org}
\and
Second Author\\
Institution2\\
First line of institution2 address\\
{\tt\small secondauthor@i2.org}
}

\maketitle
\ificcvfinal\thispagestyle{empty}\fi

\section{Experiments on ApolloSim Dataset}

\begin{table}[h]
\begin{center}
\resizebox{1\linewidth}{!}{
\begin{tabular}{c|l|cccccc}
\hline
\textbf{Scene} & \textbf{Method} & \textbf{F1(\%)$\uparrow$} & \textbf{AP(\%)$\uparrow$}  & \textbf{x err/C(m) $\downarrow$} & \textbf{x err/F(m)} $\downarrow$ & \textbf{z err/C(m) $\downarrow$} & \textbf{z err/F(m) $\downarrow$} \\
\hline
\multirow{7}{*}{Balanced Scene}     & 3DLaneNet~\cite{garnett20193d}            & 86.4          & 89.3           & 0.068          & 0.477          & 0.015          & 0.202 \\
                                    & Gen-LaneNet~\cite{guo2020gen}             & 88.1          & 90.1           & 0.061          & 0.496          & 0.012          & 0.214 \\
                                    & CLGo~\cite{liu2022learning}               & 91.9          & 94.2           & 0.061          & 0.361          & 0.029          & 0.250 \\
                                    & PersFormer~\cite{chen2022persformer}      & 92.9          & -              & 0.054          & 0.356          & 0.010          & 0.234 \\
                                    & GP~\cite{li2022reconstruct}               & 91.9          & 93.8           & 0.049          & 0.387          & \textbf{0.008} & 0.213 \\
                                    & CurveFormer \cite{bai2022curveformer}     & 95.8          & 97.3           & 0.078          & 0.326          & 0.018          & 0.219 \\
                                    & Anchor3DLane \cite{huang2023anchor3dlane} & 95.4          & 97.1           & 0.048          & 0.299          & 0.013          & 0.220 \\
                                    & BEV-LaneDet \cite{wang2022bev}            & 96.9          & -              & \textbf{0.016} & 0.242          & 0.020          & 0.216 \\
                                    & \textbf{Ours}                             & \textbf{98.2} & \textbf{99.4}  & 0.023          & \textbf{0.236} & 0.011          & \textbf{0.191} \\
\hline
\multirow{7}{*}{Rare Subset}        & 3DLaneNet~\cite{garnett20193d}            & 72.0          & 74.6           & 0.166          & 0.855          & 0.039          & 0.521 \\
                                    & Gen-LaneNet~\cite{guo2020gen}             & 78.0          & 79.0           & 0.139          & 0.903          & 0.030          & 0.539 \\
                                    & CLGo~\cite{liu2022learning}               & 86.1          & 88.3           & 0.147          & 0.735          & 0.071          & 0.609 \\
                                    & PersFormer~\cite{chen2022persformer}      & 87.5          & -              & 0.107          & 0.782          & 0.024          & 0.602 \\
                                    & GP~\cite{li2022reconstruct}               & 83.7          & 85.2           & 0.126          & 0.903          & 0.023          & 0.625 \\
                                    & CurveFormer \cite{bai2022curveformer}     & 95.6          & 97.1           & 0.182          & 0.737          & 0.039          & 0.561 \\
                                    & Anchor3DLane \cite{huang2023anchor3dlane} & 93.9          & 96.1           & \textbf{0.086} & 0.678          & 0.025          & 0.562 \\
                                    & BEV-LaneDet \cite{wang2022bev}            & 97.6          & -              & 0.031          & 0.594          & 0.040          & 0.556 \\
                                    & \textbf{Ours}                             & \textbf{98.7} & \textbf{99.7}  & 0.016          & \textbf{0.519} & \textbf{0.013} & \textbf{0.483} \\
\hline
\multirow{7}{*}{Visual Variations}  & 3D-LaneNet~\cite{garnett20193d}           & 72.5          & 74.9           & 0.115          & 0.601          & 0.032          & 0.230 \\
                                    & Gen-LaneNet~\cite{guo2020gen}             & 85.3          & 87.2           & 0.074          & 0.538          & 0.015          & 0.232 \\
                                    & CLGo~\cite{liu2022learning}               & 87.3          & 89.2           & 0.084          & 0.464          & 0.045          & 0.312 \\
                                    & PersFormer~\cite{chen2022persformer}      & 89.6          & -              & 0.074          & 0.430          & 0.015          & 0.266 \\
                                    & GP~\cite{li2022reconstruct}               & 89.9          & 92.1           & 0.060          & 0.446          & \textbf{0.011} & 0.235 \\
                                    & CurveFormer \cite{bai2022curveformer}     & 90.8          & 93.0           & 0.125          & 0.410          & 0.028          & 0.254 \\
                                    & Anchor3DLane \cite{huang2023anchor3dlane} & 92.3          & 92.6           & 0.049          & 0.363          & 0.019          & 0.242 \\
                                    & BEV-LaneDet \cite{wang2022bev}            & 95.0          & -              & 0.027          & 0.320          & 0.031          & 0.256 \\
                                    & \textbf{Ours}                             & \textbf{98.1} & \textbf{99.3}  & \textbf{0.022} & \textbf{0.239} & 0.012          & \textbf{0.213} \\
\hline   
\end{tabular}}
\caption{Comparison with state-of-the-art methods on ApolloSim dataset with three different split settings. ``C'' and ``F'' are short for close and far respectively.}
\label{tab:sota-apollo}
\end{center}
\end{table}

Apollo Synthetic dataset \cite{guo2020gen} consists of over $10$k $1080 \times 1920$ images which are built using unity 3D engine, including highway, urban, residential and downtown environments. The dataset is split into three different scenes: balanced scenes, rarely observed scenes and scenes with visual variations for evaluating algorithms from different perspectives.

Table~\ref{tab:sota-apollo} shows the experimental results under three different split settings of the ApolloSim dataset, including balanced scene, rare subset and visual variations. The experimental results show that the proposed method outperforms previous methods with large margins on F1 score and AP on all the three splits, demonstrating the superiority of our method. For example, on balanced scene, our method achieve a F1 score of $98.2$ and a AP of $99.4$, which are $2.8$ and $2.3$ points higher than those of Anchor3DLane, respectively; on rare subset, our method achieve a F1 score of $98.7$ and a AP of $99.7$, which are $4.8$ and $3.6$ points higher than those of Anchor3DLane, respectively; on visual variations, our method achieve a F1 score of $98.1$ and a AP of $99.3$, which are $5.8$ and $6.7$ points higher than those of Anchor3DLane, respectively. Our method also achieves comparable or lower x/z errors compared with previous methods, indicating that simultaneously learning lane and BEV features under supervision allows them to adjust with each other to generate more accurate predictions than the separate two-stage pipeline. It also proves that the decomposed cross-attention supervised ground-truths can achieve more accurate image-view to BEV transformation than the previous methods.


\section{Ablation Studies on Position Embeddings}

\begin{table*}[h]
	\centering 
	\resizebox{1.0\hsize}{!}{
    	\begin{tabular}{c | c c c c | c c c | c c c | c}
    		\hline
    		\multirow{2}*{\textbf{Exp.}} & \multirow{2}*{\textbf{\begin{tabular}[c]{@{}c@{}}IPM-based\\Attention\end{tabular}}} & \multirow{2}*{\textbf{\begin{tabular}[c]{@{}c@{}}Original\\Attention\end{tabular}}} & \multirow{2}*{\textbf{\begin{tabular}[c]{@{}c@{}}Decomposed\\Attention\end{tabular}}} & \multirow{2}*{\textbf{\begin{tabular}[c]{@{}c@{}}3D Pos.\\ Embedding\end{tabular}}} & \multicolumn{3}{c|}{OpenLane} & \multicolumn{3}{c|}{ONCE-3DLanes} & \multirow{2}*{\textbf{FPS}} \\
            \cline{6-11} &           &           &           &           & \textbf{F1(\%)} & \textbf{Prec.(\%)} & \textbf{Rec.(\%)}  & \textbf{F1(\%)} & \textbf{Prec.(\%)} & \textbf{Rec.(\%)} &  \\
            \hline
            1            & \ding{52} & \ding{56} & \ding{56} & \ding{56} & 57.3            & 62.1               & 53.2               & 76.8            & 80.4               & 73.5              & 25 \\
            2            & \ding{52} & \ding{56} & \ding{56} & \ding{52} & 58.6            & 63.6               & 54.3               & 77.5            & 81.3               & 74.1              & 25 \\
            3            & \ding{56} & \ding{52} & \ding{56} & \ding{56} & 56.1            & 57.2               & 54.9               & 75.9            & 77.3               & 74.5              & 17 \\
            4            & \ding{56} & \ding{52} & \ding{56} & \ding{52} & 59.5            & 63.4               & 56.2               & 78.2            & 81.5               & 75.2              & 17 \\
            5            & \ding{56} & \ding{56} & \ding{52} & \ding{56} & 59.7            & 65.8               & 54.6               & 78.7            & 83.5               & 74.4              & 45 \\
            6            & \ding{56} & \ding{56} & \ding{52} & \ding{52} & \bf{62.8}       & \bf{67.8}          & \bf{58.5}          & \bf{80.5}       & \bf{84.7}          & \bf{76.7}         & 45 \\
    		\hline
    	\end{tabular}
 	}
	\caption{Ablation studies about image-view and BEV position embeddings on OpenLane \cite{chen2022persformer} and ONCE-3DLanes \cite{yan2022once} datasets.}
	\label{Tab4}
	\vspace{-4mm}
\end{table*}

\begin{figure}[h]
\centering
\includegraphics[width=0.6\linewidth]{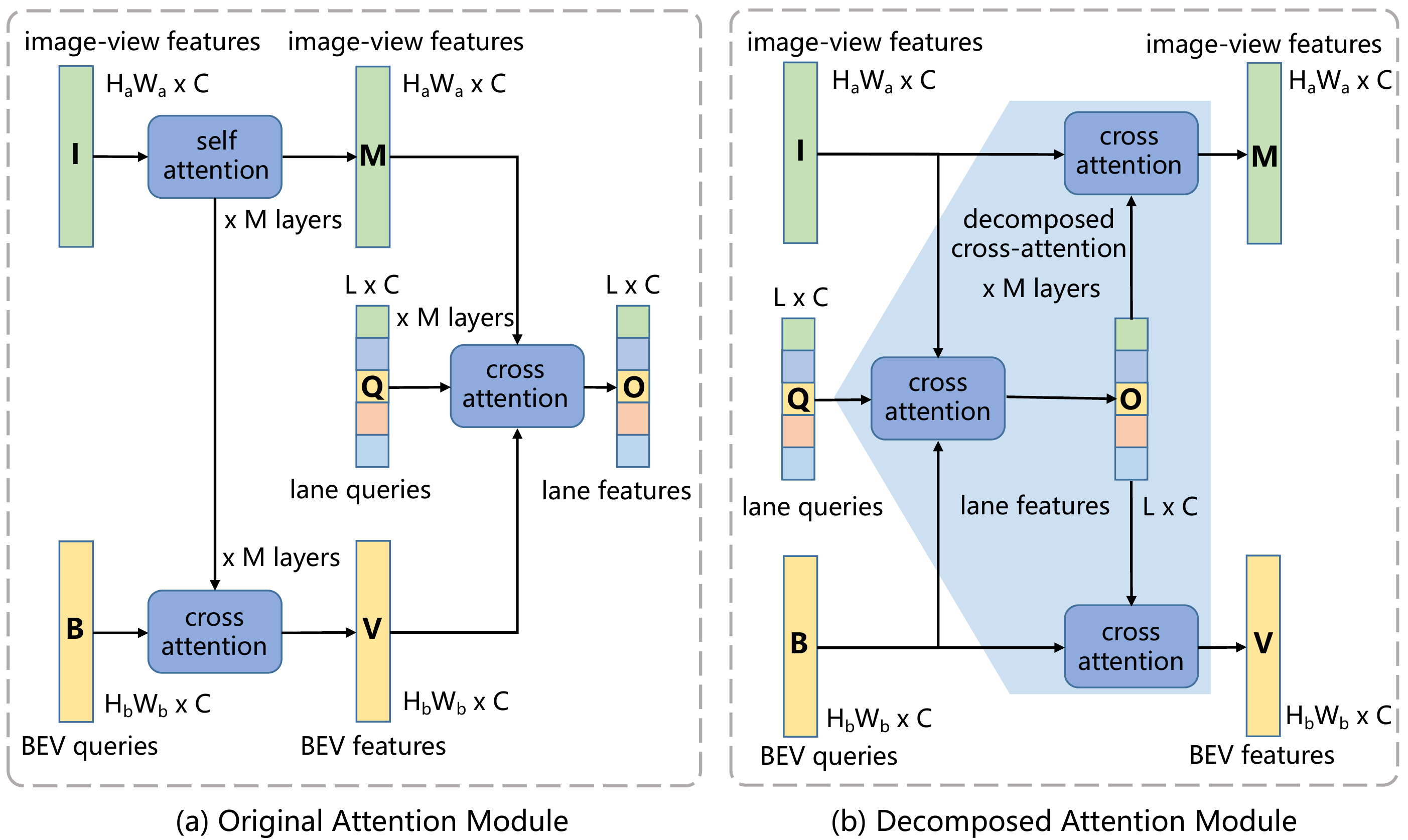}
\caption{The comparison of learning lane and BEV representations between the original attention module and the decomposed attention module.}
\label{Fig:frame_comp}
\end{figure}

We compare the results with and without the proposed position embedding. For the results without the proposed position embedding, we apply the original position embedding used in \cite{vaswani2017attention}, where sine/cosine position embedding is used in the transformer encoder and learnable position embedding is used in the transformer decoder. We validate the effectiveness of the proposed position embedding in three different view transformation (image-view feature map to bird's-eye-view feature map) methods, including the IPM-based attention, original attention and decomposed attention. The IPM-based attention is used in PersFormer \cite{chen2022persformer} for view transformation, where the reference points of the deformable transformer are computed with inverse perspective mapping (IPM). The original attention and the decomposed attention are shown in Figure \ref{Fig:frame_comp} (a) and (b), respectively. The former transforms the image-view feature map into bird's-eye-view (BEV) directly with a cross-attention mechanism proposed in \cite{vaswani2017attention}. The latter decomposes the cross-attention between image-view and BEV features into the one between image-view and lane features, and the one between lane and BEV features, both of which are supervised with ground-truth lane lines. As for the implementation details of the proposed position embedding in Section 3.3, the number of discretized depth bins $D$ is set to $50$, covering a depth range of $\left[0, 100\right] m$; the number of discretized height bins $Z$ is set to $50$, covering a height range of $\left[-5, 5\right] m$.

As shown in Table \ref{Tab4}, the results with the proposed position embedding are better than the results without it. Specifically, for the IPM-based attention, the F1 scores of the proposed position embedding are $1.3$ and $0.7$ points higher than those of the original position embedding on OpenLane and ONCE-3DLanes, respectively; for the original attention, the F1 scores of the proposed position embedding are $3.4$ and $2.3$ points higher than those of the original position embedding on OpenLane and ONCE-3DLanes, respectively; for the decomposed attention, the F1 scores of the proposed position embedding are $3.1$ and$1.8$ points higher than those of the original position embedding on OpenLane and ONCE-3DLanes, respectively. These results demonstrate that the proposed position embedding can effectively improve the performance of view transformation. This is because the 3D position embedding can accurately align image-view and BEV features by incorporating 3D coordinate information into both of them according to the predicted depth and height distributions. We also find that the improvements for the original attention and the decomposed attention are higher than the improvement for the IPM-based attention. This is because the view transformation of the IPM-based attention is mainly determined by the fixed reference points computed with IPM, making the effect of the proposed position embedding limited.

{\small
\bibliographystyle{ieee_fullname}
\bibliography{egbib}
}